\documentclass[sigconf]{acmart}
\AtBeginDocument{%
  \providecommand\BibTeX{{%
    \normalfont B\kern-0.5em{\scshape i\kern-0.25em b}\kern-0.8em\TeX}}}

\acmPrice{15.00}
\acmISBN{978-1-4503-XXXX-X/18/06}

\usepackage{amsfonts}
\usepackage{multirow}
\usepackage{hhline}

\begin{document}

\title{Debiasing Gender Bias in Information Retrieval Models}

\author{Dhanasekar Sundararaman}

\email{dhanasekar.sundararaman@duke.edu}
\affiliation{%
  \institution{Duke University}
  \city{Durham}
  \state{North Carolina}
  \country{USA}
  \postcode{43017-6221}
}

\author{Vivek Subramanian}

\email{viveksub@amazon.com}
\affiliation{%
  \institution{Amazon Alexa AI}
  \city{Boston}
  \state{Massachusetts}
  \country{USA}
}

\setcopyright{none}

\settopmatter{printacmref=false} 

\renewcommand{\shortauthors}{Sundararaman and Subramanian}
\renewcommand\footnotetextcopyrightpermission[1]{}
\pagestyle{plain}

\begin{abstract}
    Biases in culture, gender, ethnicity, etc. have existed for decades and have affected many areas of human social interaction. These biases have been shown to impact machine learning (ML) models, and for natural language processing (NLP), this can have severe consequences for downstream tasks. Mitigating gender bias in information retrieval (IR) is important to avoid propagating stereotypes. In this work, we employ a dataset consisting of two components: (1) relevance of a document to a query and (2) ``gender'' of a document, in which pronouns are replaced by male, female, and neutral conjugations. We definitively show that pre-trained models for IR do not perform well in zero-shot retrieval tasks when full fine-tuning of a large pre-trained BERT encoder is performed and that lightweight fine-tuning performed with adapter networks improves zero-shot retrieval performance almost by 20\% over baseline. We also illustrate that pre-trained models have gender biases that result in retrieved articles tending to be more often male than female. We overcome this by introducing a debiasing technique that penalizes the model when it prefers male over female, resulting in an effective model that retrieves articles in a balanced fashion across genders.
\end{abstract}



\maketitle



\section{Introduction}
Differences in culture, gender, ethnicity, etc. naturally lead to conscious and unconscious bias among humans \cite{silberg2019notes}. These biases have existed for decades and have affected many areas of human social interaction such as admissions to higher institutions, job selection, and financial compensation. Such human biases are incorporated into artificial intelligence (AI) models, which are trained on examples generated by humans. It is critical to identify and mitigate these biases as AIs become more ubiquitous in day to day tasks.

Biases have been shown to impact all types of machine learning (ML) models, including computer vision (CV) and natural language processing (NLP) \cite{sundararaman2019syntax, sundararaman2021syntactic} . For instance, algorithms for facial recognition have been shown to disproportionately identify black and hispanic males as criminals among a diverse set of images \cite{richardson2019dirty}. In addition, it was shown that algorithms developed by leaders in facial recognition more often misclassified black women than white males \cite{lohr2018facial}. In the NLP community, embeddings trained on large corpora of data \cite{shen2019learning} were more likely to associate African American names with unpleasant words \cite{caliskan2017semantics}. Generative models for images from text, when queried to generate images of doctors or engineers, typically generate male rather than female individuals \cite{dev2019attenuating}. These biases can feed into various downstream NLP tasks \cite{sundararaman2020methods, sundararaman2022number} including summarization, generation \cite{subramanian2021lexical}, and information retrieval.


Information retrieval (IR) is the task of identifying the document in a database which is most relevant to a given query. There are two main tasks in IR: (1) representing queries and documents and (2) scoring relevance between the two. With the advent of deep learning, task (1) is typically accomplished by utilizing large, pretrained language models such as BERT to generate embeddings for sequences of tokens \cite{devlin2018bert}. Then, scoring relevance can be accomplished by computing some similarity metric (e.g., cosine similarity \cite{singhal2001modern}) between query and document embeddings. One of the main shortcomings of BERT is that it is trained on two large corpora of text -- BooksCorpus (800M words) and Wikipedia (2500M words) \cite{devlin2018bert} -- which are known to suffer from gender bias \cite{bhardwaj2021investigating}. The bias present in BERT could be realized through its performance on a number of downstream tasks ranging from classification to tagging.

Eliminating gender bias in IR is important since biased retrieval can result in unwanted propagation of negative gender stereotypes. For instance, as shown in \cite{garg2018word}, word embeddings derived from over 100 years of text data evolve with occupational patterns observed among males and females. Utilizing these embeddings, which associate gender with occupational roles that may not be relevant in today's context, can bias an IR model. Models pretrained using such embeddings tend to be biased towards a (gender, occupation) pair even assuming there is an ideal dataset in which the frequency of each gender  label is equal. For example, an IR model could reinforce stereotypes of traditional female roles such as secretary, nurse, or housekeeper while diminishing the roles of women who are engineers, mechanics, or carpenters \cite{garg2018word}.

\begin{table*}[t]
\caption{Zero-shot retrieval results using full fine-tuning.}
\begin{tabular}{c|c|c|c|c|c|c|c|c}
\hline
Train/Test &
  \begin{tabular}[c]{@{}c@{}}Sex \&\\ Relationship\end{tabular} &
  Career &
  \begin{tabular}[c]{@{}c@{}}Child\\ Care\end{tabular} &
  Appearance &
  \begin{tabular}[c]{@{}c@{}}Cognitive \\ Capabilities\end{tabular} &
  \begin{tabular}[c]{@{}c@{}}Domestic\\ Work\end{tabular} &
  \begin{tabular}[c]{@{}c@{}}Physical\\ Capabilities\end{tabular} &
  \begin{tabular}[c]{@{}c@{}}Average*\\ \end{tabular} \\ \hline
Sex \& Relationships   & 0.9565                & 0.4833                & 0.4167                & 0.5238                & 0.5139                & 0.5000                & 0.4912  & 0.4882 \\ \hline
Career                 & 0.5290                & 0.8250                & 0.6548                & 0.5476                & 0.6111                & 0.4889                & 0.4912 & 0.5538\\ \hline
Child Care             & 0.5942                & 0.5250                & 1.0000                & 0.5119                & 0.5000                & 0.4111                & 0.5526 & 0.5158 \\ \hline
Appearance             & 0.4783                & 0.4250                & 0.5476                & 1.0000                & 0.4028                & 0.5444                & 0.5263 & 0.4874\\ \hline
Cognitive Capabilities & 0.5217                & 0.4917                & 0.5000                & 0.4762                & 0.8333                & 0.5111                & 0.4737  & 0.4957\\ \hline
Domestic Work          & 0.5435                & 0.5000                & 0.4643                & 0.6071                & 0.5972                & 0.9000                & 0.4737 & 0.531\\ \hline
Physical Capabilities  & 0.4855                & 0.4750                & 0.5476                & 0.5476                & 0.5833                & 0.5333                & 1.0000 & 0.5287\\ \hline

\end{tabular}
\label{tab:fullfinetuning}
\end{table*}

Very few methods have been proposed to reduce gender bias in IR. In image retrieval, it has been shown that women are less likely to be identified as working at a computer \cite{hendricks2018women}. This bias was mitigated by adding two complementary loss functions which both confuse the model when gender information is not present in an image and make the model more confident when it is. Similarly, in text retrieval, methods have focused on balancing the number of examples associated with each gender, but this does not necessarily eliminate more complex biases at a model-level that correlate gender with certain features \cite{bigdeli2021exploration}.

In addition, with the plethora of NLP/IR datasets that are being collected and released, it becomes increasingly unlikely that the distribution of the data on which a model must be evaluated matches that of the data used for pretraining. This motivates the need for zero-shot learning and retrieval methods, which allow a model to generalize to unseen dataset distributions. In addition, evaluating a pretrained model on a zero-shot basis can help us to learn more about the bias of a model since the training and test distributions are no longer connected.

In our experiments, we employ the \texttt{\href{https://github.com/KlaraKrieg/GrepBiasIR}{GrepBiasIR}} dataset \cite{krieg2022grep}, which consists of seven categories of query-document pairs. (More details provided in section \ref{sec:datasets}.) For a very small dataset, BERT, being a model with millions of parameters, is likely to overfit with full fine-tuning. Thus, there is a need for lightweight fine-tuning that effectively uses BERT parameters while also generalizing well. Adapter networks, single layer networks that can be introduced into a complex model, achieve this by allowing us to freeze BERT parameters while updating adapter layers. We show that this form of fine-tuning performs better on \texttt{GrepBiasIR} than the full fine-tuning approach, producing an average improvement of over 20\%.

To summarize our contributions, in this paper, we analyze the information retrieval capability of a BERT model over several categories of a bias IR dataset. By fine-tuning on one category and testing on the others, we are able to show the bias of BERT towards different genders at a more granular level. It is imperative that fine-tuning takes place only on a single category of data because we want to show that our method generalizes well to classes outside the one on which the model is fine-tuned. We also introduce a regularization-based debiasing technique for mitigating the aforementioned gender bias. Our methods can be used to debias BERT for a wide variety of applications, including but not limited to IR.

\begin{table*}[t]
\caption{Zero-shot retrieval using adapter-based fine-tuning.}
\begin{tabular}{c|c|c|c|c|c|c|c|c}
\hline
Train/Test &
  \begin{tabular}[c]{@{}c@{}}Sex \&\\ Relationships\end{tabular} &
  Career &
  \begin{tabular}[c]{@{}c@{}}Child\\ Care\end{tabular} &
  Appearance &
  \begin{tabular}[c]{@{}c@{}}Cognitive \\ Capabilities\end{tabular} &
  \begin{tabular}[c]{@{}c@{}}Domestic\\ Work\end{tabular} &
  \begin{tabular}[c]{@{}c@{}}Physical\\ Capabilities\end{tabular} &
  \begin{tabular}[c]{@{}c@{}}Average*\\ \end{tabular} \\ \hline
Sex \& Relationships          & 0.9565                & 0.6250                & 0.5357                & 0.6548                & 0.8194                & 0.7222                & 0.7193  & 0.6794              \\ \hline
Career             & 0.8188                & 0.8917                & 0.5714                & 0.7024                & 0.7500                & 0.7333                & 0.7018 & 0.7130 \\ \hline
Child Care             & 0.7536                & 0.6833                & 0.9048                & 0.7500                & 0.6944                & 0.7222                & 0.7105 & 0.7190 \\ \hline
Appearance             & 0.7826                & 0.6167                & 0.5714                & 0.9286                & 0.7917                & 0.7222                & 0.7105 & 0.6992 \\ \hline
Cognitive Capabilities & 0.8188                & 0.6083                & 0.5357                & 0.6429                & 0.9306                & 0.6778                & 0.6842 & 0.6613 \\ \hline
Domestic Work          & 0.7899                & 0.6167                & 0.5476                & 0.6667                & 0.7917                & 0.9000                & 0.7018 & 0.6857\\ \hline
Physical Capabilities  & 0.8188                & 0.5750                & 0.6071                & 0.7143                & 0.7778                & 0.6889                & 0.8947 & 0.6970\\ \hline
\end{tabular}
\label{tab:adapterfinetuning}
\end{table*}

\begin{table*}[t]
\caption{Fraction of male, female, and neutral articles with highest logits. Model is biased towards male.}
\resizebox{\textwidth}{!}{%
\begin{tabular}{c|ccc|ccc|ccc|ccc|ccc|ccc|ccc|ccc}

  \hline
 Train/Test & \multicolumn{3}{c|}{\begin{tabular}[c]{@{}c@{}}Sex \&\\ Relationships\end{tabular} } & \multicolumn{3}{c|}{Career} & \multicolumn{3}{c|}{\begin{tabular}[c]{@{}c@{}}Child\\ Care\end{tabular}} &
 \multicolumn{3}{c|}{Appearance} &
 \multicolumn{3}{c|}{\begin{tabular}[c]{@{}c@{}}Cognitive \\ Capabilities\end{tabular} } &
 \multicolumn{3}{c|}{\begin{tabular}[c]{@{}c@{}}Domestic \\ Work\end{tabular}} &
 \multicolumn{3}{c|}{\begin{tabular}[c]{@{}c@{}}Physical \\ Capabilities\end{tabular} } &
 \multicolumn{3}{c}{Average}\\ \cline{2-25}

& M & F & N 
& M & F & N
& M & F & N
& M & F & N
& M & F & N
& M & F & N
& M & F & N
& M & F & N\\ \hline
                    
Sex \& Relationships & 0.13 & 0.18 & 0.69 
                    & 0.43 & 0    & 0.57        
                    & 0.45 & 0    & 0.55
                    & 0.25 & 0.33 & 0.42
                    & 0.14 & 0    & 0.86
                    & 0.33 & 0.33 & 0.33
                    & 0.46 & 0.23 & 0.31
                    & 0.32 & 0.2 & 0.58\\ \hline

Career & 0.14 & 0.14 & 0.72 
                    & 0.50  & 0.06 & 0.44        
                    & 0.40  & 0.20  & 0.40
                    & 0.33 & 0.17 & 0.50
                    & 0    & 0    & 1
                    & 0.40  & 0.20  & 0.40
                    & 0.36 & 0.18 & 0.46
                    & 0.31 & 0.13 & 0.56
                    \\ \hline
Child Care & 0.08 & 0.25 & 0.67 
                    & 0.15 & 0.31    & 0.54        
                    & 0.25 & 0.17    & 0.58
                    & 0.08 & 0.50    & 0.42
                    & 0    & 0.14    & 0.86
                    & 0.40  & 0.20     & 0.40
                    & 0.21 & 0.50    & 0.29
                    & 0.16 & 0.3    & 0.53\\ \hline
Appearance & 0.14 & 0.22 & 0.64 
                    & 0.23 & 0.23 & 0.54        
                    & 0.50  & 0    & 0.50
                    & 0.25 & 0.25 & 0.50
                    & 0    & 0    & 1
                    & 0    & 0.20  & 0.80
                    & 0.36 & 0.45 & 0.19 
                    & 0.21 & 0.19 & 0.6\\ \hline
Cognitive Capabilities & 0.13 & 0.20 & 0.67 
                    & 0.33 & 0.20  & 0.47        
                    & 0.42 & 0    & 0.58
                    & 0.25 & 0.17 & 0.58
                    & 0.25 & 0    & 0.75
                    & 0.17 & 0    & 0.83
                    & 0.15 & 0.23 & 0.62
                    & 0.25 & 0.11 & 0.64\\ \hline
Domestic Work & 0.13 & 0.20 & 0.67 
                    & 0.50 & 0.21   & 0.29        
                    & 0.50 & 0      & 0.50
                    & 0.25 & 0.25  & 0.50
                    & 0    & 0     & 1
                    & 0.14 & 0.43  & 0.43
                    & 0.46 & 0.46  & 0.08 
                    & 0.28 & 0.22  & 0.5\\ \hline
Physical Capabilities & 0.20 & 0.13 & 0.67 
                    & 0.31 & 0.31    & 0.38        
                    & 0.40 & 0    & 0.60
                    & 0.25 & 0.25 & 0.50
                    & 0.29 & 0    & 0.71
                    & 0.40 & 0.20 & 0.40
                    & 0.31 & 0.31 & 0.38
                    & 0.30 & 0.18 & 0.52\\ \hline
\end{tabular} %
}
\label{tab:bias}
\end{table*}

\section{Related Works}
The topic of gender bias in ML models is highly debated and studied. While there have been some significant contributions in the past, not a lot of emphasis has been shown particularly on retrieval models, let alone zero-shot transfer. Works such as \cite{bolukbasi2016man} identified the gender bias problem in ML models, showing the intrinsic bias in word embeddings that could potentially propogate to numerous downstream tasks. \cite{zhao-etal-2018-learning} proposed methods for gender-neutral static embeddings by preserving gender information in certain dimensions of a word embedding while others are free from gender information.

In retrieval, \cite{rekabsaz2020neural} has methodologically analyzed gender bias in prominent retrieval models. They found that all models exhibited strong bias towards male and that the bias is pronounced in neural models, which corroborates our findings. The neural models exhibit significant bias because of the pre-trained embeddings trained from biased corpora. To the best of our knowledge, we are the first to identify gender bias in retrieval models on a zero-shot basis, and to propose methods to mitigate them while not compromising retrieval performance.

\section{Preliminaries}
\subsection{Datasets}
\label{sec:datasets}
For our experiments, we employ the \texttt{\href{https://github.com/KlaraKrieg/GrepBiasIR}{GrepBiasIR}} dataset \cite{krieg2022grep}, which consists of 819 query, document title, and document content tuples divided roughly evenly into seven categories: Appearance, Career, Child Care, Cognitive Capability, Domestic Work, Physical Capability, and Sex and Relationships. Queries in each category are each paired with two documents: one which is relevant to the query and one which is not. In addition, three versions of each document have been created by substituting pronouns with their male ($M$), female ($F$), and neutral ($N$) forms. This allows us to assess retrieval performance, bias of a pretrained model, and bias correction after fine-tuning.

\subsection{Training}

For full fine-tuning, we train for 2 epochs using an AdamW Optimizer with learning rate $2\times 10^{-5}$ and batch size $B$ of $8$. This is the maximum number of epochs on which the model can be trained without validation accuracy saturating. The model is trained with cross-entropy loss. For adapter fine-tuning, we train for 8 epochs with learning rate $1\times 10^{-4}$ using the same optimizer and batch size.

\section{Methods}


\subsection{Adapter-based fine-tuning}
Our dataset consists of $N$ query-document pairs $x_i \: \forall \: i \in \{1, \ldots, N\}$. To classify a document's relevancy to the query, we first employ a BERT encoder which maps each data point to an embedding $\mathbf{x}_i \in \mathbb{R}^D$, where $D=768$, via the BERT \texttt{CLS} token \cite{devlin2018bert}. This query-document embedding is then directly passed through a binary classifier, which is the adapter layer in our model, to obtain a score for the data point. Mathematically, we have:
\begin{align}
    y_i = \sigma(\mathbf{A}\mathbf{x}_i) = \sigma(z_i)
\end{align}
where $\mathbf{A} \in \mathbb{R}^{D}$ is a learned vector of weights, $z_i \in \mathbb{R}$ are logits for relevancy scores, and $\sigma(\cdot) : \mathbb{R} \mapsto (0, 1)$ is the logistic sigmoid activation function: $\sigma(z) = \frac{1}{1+\exp(-z)}$. We fine-tune these weights on a per-category basis to obtain retrieval scores for all other categories. For each data point, the loss is the binary cross-entropy loss:
\begin{align}
    \mathcal{L}_i = -t_i\log(y_i) - (1- t_i)\log(1-y_i)
\end{align}
where $t_i \in \{0, 1\}$ is the target label for data point $x_i$. This is averaged over a batch, and weights in the output adapter layer are updated by gradient descent. Importantly, the weights of the BERT encoder remain fixed throughout the fine-tuning process, so each update is roughly $\mathcal{O}(D)$.

\subsection{Mitigating gender bias}
While the above method is an effective way to build a classifier for determining relevance of a document to a query, in practice, we find that documents whose pronouns are male-gendered tend to be more relevant to the same query than the corresponding female-gendered documents. This reflects the well-known gender bias in the BERT model.

To mitigate this bias, we employ a regularization term which penalizes for preference of one gender over another. In each batch (indexed by $j$), we randomly sample without replacement pairs (indexed by $k$) of data points $(x_k, x_k')$ whose documents are of different genders $(g_k, g_k')$, where $g_k, g_k' \in \{M, F, N\}$. We then compute the sums of squared differences between relevancy logits $(z_k, z_k')$ of query-document pairs within a batch that are of different genders. These are averaged across all samples in a batch to compute the regularization term $\mathcal{R}_j$ for that batch. Each pair of differences is weighted by an $\alpha_k \in \mathbb{R}^+$, whose value depends on the genders of the documents involved in the pair. Thus, there are a total of ${G \choose 2} = 3$ unique $\alpha_k$s, where $G=3$ is the number of genders, which are tuned as hyperparameters by grid search over $0$ to $2$ in increments of $0.25$. Mathematically, we have:
\begin{align}
    \mathcal{R}_j = \frac{1}{N_j}\sum_{\{k \: : \: g_k \neq g_k'\}}\alpha_k(\mathcal{G}(x_k), \mathcal{G}(x_k'))(z_k - z_k')^2
\end{align}
where $N_j$ is the total number of pairs in batch $j$ and $\mathcal{G}(x_k) : x_k \mapsto g_k$. Thus, the total loss for a batch becomes:
\begin{align}
    \mathcal{L}_j = \frac{1}{B}\sum_{i=1}^{B}\mathcal{L}_i + \mathcal{R}_j
\end{align}

\section{Results and Discussion}







\begin{table}[t]
\caption{Top words in each category based on TF-IDF.}
\begin{tabular}{c|c}
\hline
Category               & Top-3 TF-IDF words       \\ \hhline{=|=}
Sex and Relationships  & love (0.18), emotional (0.06), friend (0.06) \\ \hline
Career                 & job (0.10), career (0.08), home (0.08) \\ \hline
Child Care             & baby (0.34), parents (0.11), children (0.08) \\ \hline
Appearance             & beauty (0.10), hair (0.10), attractive (0.06) \\ \hline
Cognitive Capabilities & iq (0.09), intelligence (0.06), gifted (0.06) \\ \hline
Domestic Work          & home (0.14), cleaning (0.11), house (0.10) \\ \hline
Physical Capabilities  & athletes (0.17), fat (0.11), weight (0.10) \\ \hline
\end{tabular}
\label{tab:wordfrequencies}
\end{table}

\begin{table*}[t]
\caption{Zero-shot retrieval after de-biasing.}
\begin{tabular}{c|c|c|c|c|c|c|c|c}
\hline
Train/Test &
  \begin{tabular}[c]{@{}c@{}}Sex \&\\ Relationships\end{tabular} &
  Career &
  \begin{tabular}[c]{@{}c@{}}Child\\ Care\end{tabular} &
  Appearance &
  \begin{tabular}[c]{@{}c@{}}Cognitive \\ Capabilities\end{tabular} &
  \begin{tabular}[c]{@{}c@{}}Domestic\\ Work\end{tabular} &
  \begin{tabular}[c]{@{}c@{}}Physical\\ Capabilities\end{tabular} &
  \begin{tabular}[c]{@{}c@{}}Average*\\ \end{tabular} \\ \hline
Sex \& Relationships          & 0.7681                & 0.5750                & 0.5238                & 0.6310                & 0.7639                & 0.6778                & 0.6579  & 0.6382              \\ \hline
Career             & 0.6594                & 0.6416                & 0.6785                & 0.6904                & 0.7083                & 0.6888                & 0.6842 & 0.6849 \\ \hline
Child Care             & 0.7391                & 0.6167                & 0.6667                & 0.7500                & 0.7500                & 0.7222                & 0.6842 & 0.7103 \\ \hline
Appearance             & 0.7681                & 0.6250                & 0.6429                & 0.7857                & 0.7917                & 0.7222                & 0.7105 & 0.7100 \\ \hline
Cognitive Capabilities & 0.6231                & 0.4833                & 0.5595                & 0.5833                & 0.8472                & 0.5888                & 0.6754 & 0.5856 \\ \hline
Domestic Work          & 0.7899               & 0.6167                & 0.5476                & 0.6667                & 0.7917                & 0.9000                & 0.7018 & 0.6857\\ \hline
Physical Capabilities  & 0.7463               & 0.6083                & 0.6309                & 0.7261                & 0.8333                & 0.7555                & 0.7192 & 0.7167\\ \hline
\end{tabular}
\label{tab:debiasing}
\end{table*}

\begin{table}[t]
\caption{Percentage difference between average male and female, before and after de-biasing.}
\resizebox{\columnwidth}{!}{%
\begin{tabular}{l|lll|lll|ll}
\hline
 & \multicolumn{3}{c|}{Average Before} & \multicolumn{3}{c|}{Average After} & \multicolumn{2}{c}{$|\mbox{M}-\mbox{F}|$} \\ \cline{2-9} 
 & M & F & N & M & F & N & Before & After \\ \hline
Sex \& Relationships & 0.32 & 0.2 & 0.58 & 0.21 & 0.27 & 0.52 & 0.12 & \textbf{0.06} \\ \hline
Career & 0.31 & 0.13 & 0.56 & 0.28 & 0.22 & 0.5 & 0.18 & \textbf{0.06} \\ \hline
Child Care & 0.16 & 0.3 & 0.53 & 0.22 & 0.26 & 0.52 & 0.14 & \textbf{0.04} \\ \hline
Appearance & 0.21 & 0.19 & 0.6 & 0.21 & 0.24 & 0.55 & 0.02 & 0.03 \\ \hline
Cognitive Capabilities & 0.25 & 0.11 & 0.64 & 0.27 & 0.36 & 0.37 & 0.14 & \textbf{0.09} \\ \hline
Domestic Work & 0.28 & 0.22 & 0.5 & 0.28 & 0.22 & 0.5 & 0.06 & 0.06 \\ \hline
Physical Capabilities & 0.3 & 0.18 & 0.52 & 0.31 & 0.25 & 0.44 & 0.12 & \textbf{0.06} \\
\hline
\end{tabular}}
\label{tab:biascomparison}
\end{table}

Table \ref{tab:fullfinetuning} shows the retrieval accuracy obtained from our full fine-tuning experiments on each of the seven categories. (In all tables, columns denoted Average* indicate that the average was taken across a row without the diagonal element, to quantify zero-shot performance.) Baseline accuracy on each class is roughly chance (50\%) before fine-tuning. After full fine-tuning of BERT, in which all layers of BERT are fine-tuned, we find that the retrieval performance on data from the same class as that in which fine-tuning was performed greatly improves (retrieval accuracy of $0.9307\pm 0.0782$). For the remaining classes, we find that zero-shot retrieval performance still hovers around chance (average of averages for each class $0.5144 \pm 0.0252$), illustrating the poor generalization performance of full fine-tuning.

Table \ref{tab:adapterfinetuning} shows that fine-tuning using an adapter layer significantly boosts performance compared to full fine-tuning. Specifically, average performance on the same class as the fine-tuning data is $0.9153 \pm 0.0239$. Although the average performance is slightly lower, the error bar is much tighter, likely because there are fewer parameters over which additional uncertainty is introduced. Average of averages for fine-tuning with an adapter model on each class is $0.6935 \pm 0.0199$, an improvement of almost 20\%, indicating much greater zero-shot generalizability after this type of fine-tuning.

Table \ref{tab:bias} shows a detailed breakdown by gender of the results from Table \ref{tab:adapterfinetuning}. Specifically, for each class on which we fine-tuned, we show, for the correctly retrieved examples, the fraction for which the gender label with the highest relevancy logit value was $M$, $F$, or $N$. On average, we find that the absolute difference between $M$ and $F$ fractions is $11.14 \pm 5.39$\%, exemplifying the strong gender bias of the BERT model. Note that examples for which the model retrieved the male and female versions of a document with equal probability were discarded to avoid diluting the bias with examples for which the model was equally confident about both genders.

We find that, on average, in six out of the seven categories, the male version of a document is preferred over a female version, with the female only being preferred when fine-tuning is done on Child Care. In two of the cases (Career and Cognitive Capabilities), the male version is preferred by more than a factor of two. Looking closely at each row of the table, we find that, after fine-tuning on Career, the model is more likely to predict the male version of a document when classifying in all other categories, except Sex and Relationships, where the male and female genders are equally represented. Similarly, when fine-tuning on Child Care, the model is significantly more likely to predict the female version of a document in all categories but Domestic Work, in which the male version is preferred by a factor of two (domestic work generally is biased towards the male gender across categories of documents). This suggests that, even on a balanced dataset in which male and female documents are represented with equal frequency, the \{Sex and Relationships, Career\} and Child Care categories are respectively biased towards the male and female gendered documents.

Table \ref{tab:wordfrequencies} shows the words in each category with the highest term frequency-inverse document frequency (TF-IDF) scores \cite{rajaraman2011mining}. These results, along with the bias results shown in Table \ref{tab:bias}, illustrate that salient words like \{love, emotional, friend\} (for Sex and Relationships) or \{job, career, home\} (for Career) are incorrectly associated to the male gender, even though they are inherently gender-neutral. Likewise, \{baby, parents, children\} (for Child Care) are incorrectly associated to the female gender. These words lead to bias that potentially drives the model to prefer one gender over another.

Finally, we investigate the performance of our model after employing our debiasing technique. Reducing the gender gap in zero-shot conditions is challenging because we only have access to the validation accuracy on the category on which the model is fine-tuned.
Tables \ref{tab:debiasing} and \ref{tab:biascomparison} show the updated retrieval metrics after regularizing for gender difference. Our experiments showed that with strong regularization ($\alpha_k >> 2 \: \forall \: k$), we were able to bring the gender bias close to zero; however, this was at the cost of a severe degradation of retrieval performance. Thus, there is a need to balance drop in retrieval performance with reduction in gender bias. With this in mind, we fine-tuned the $\alpha_k$'s such that the retrieval performance did not drop by more than 2-5\% while the difference between male and female bias was brought down to an average of 5.7\%. Performance after fine-tuning on Cognitive Capabilities dropped the most, indicating that words in this category were especially skewed towards male.

\section{Conclusion}
Gender bias is a well-known, prevalent issue in many areas of NLP, including IR. Mitigating gender bias is an especially difficult problem in the context of zero-shot retrieval. In this paper, we demonstrate that employing adapter-based fine-tuning yields significantly better zero-shot retrieval performance compared to full fine-tuning of a BERT encoder. To reduce gender bias, we further introduced a debiasing regularization method, and show that we gain about 6\% reduction in gender bias at the cost of 2-5\% reduction in retrieval performance. Natural extensions of this work include developing methods for addressing multiple forms of bias that can be present in a single model, and in employing latent variable methods to intuit this bias when labeled data is not available.

\bibliographystyle{ACM-Reference-Format}
\bibliography{sample-base}


\end{document}